\documentclass[sigconf]{acmart}
\AtBeginDocument{%
  }

\newcommand{\KK}[1]{{}}
\newcommand{\YM}[1]{{}}
\newcommand{\GC}[1]{}
\newcommand{\MB}[1]{}

\setcopyright{acmlicensed}
\copyrightyear{2025}
\acmYear{2025}
\acmDOI{XXXXXXX.XXXXXXX}
\acmConference[Conference acronym 'XX]{Make sure to enter the correct
  conference title from your rights confirmation email}{June 03--05,
  2018}{Woodstock, NY}
\acmISBN{978-1-4503-XXXX-X/2018/06}

\begin{document}

%%
%% The "title" command has an optional parameter,
%% allowing the author to define a "short title" to be used in page headers.
\title{Large Language Models for Geolocation Extraction in Humanitarian Crisis Response}

%%
%% The "author" command and its associated commands are used to define
%% the authors and their affiliations.
%% Of note is the shared affiliation of the first two authors, and the
%% "authornote" and "authornotemark" commands
%% used to denote shared contribution to the research.

\author{Gianmarco Cafferata}
\email{gcafferata@udesa.edu.ar}
\orcid{0009-0006-6606-8448}
\affiliation{%
  \institution{Universidad de San Andrés}
  \city{Victoria}
  \state{Buenos Aires}
  \country{Argentina}
}

\author{Tiziano Demarco}
\email{demarcot@udesa.edu.ar}
\orcid{0009-0007-4423-596X}
\affiliation{%
  \institution{Universidad de San Andrés}
  \city{Victoria}
  \state{Buenos Aires}
  \country{Argentina}
}

\author{Kyriaki Kalimeri}
\email{kyriaki.kalimeri@isi.it}
\orcid{0000-0001-8068-5916}
\affiliation{%
  \institution{ISI Foundation}
  \city{Turin}
  \country{Italy}
}
\author{Yelena Mejova}
\email{yelena.mejova@isi.it}
\orcid{0000-0001-5560-4109}
\affiliation{%
\institution{ISI Foundation}
  \city{Turin}
  \country{Italy}
}

\author{Mariano G. Beiró}
\email{mbeiro@udesa.edu.ar}
\orcid{0000-0002-5474-0309}
\affiliation{%
  \institution{Universidad de San Andrés}
  \city{Victoria}
  \state{Buenos Aires}
  \country{Argentina}
}
\affiliation{%
  \institution{CONICET}
  \city{Buenos Aires}
  \country{Argentina}
}

%%
%% By default, the full list of authors will be used in the page
%% headers. Often, this list is too long, and will overlap
%% other information printed in the page headers. This command allows
%% the author to define a more concise list
%% of authors' names for this purpose.

%\renewcommand{\shortauthors}{Cafferata et al.}
\renewcommand{\shortauthors}{Gianmarco Cafferata, Tiziano Demarco, Kyriaki Kalimeri, Yelena Mejova, and Mariano G. Beiró}

%%
%% The abstract is a short summary of the work to be presented in the
%% article.
\begin{abstract}
Humanitarian crises demand timely and accurate geographic information to inform effective response efforts. Yet, automated systems that extract locations from text often reproduce existing geographic and socioeconomic biases, leading to uneven visibility of crisis-affected regions. This paper investigates whether Large Language Models (LLMs) can address these geographic disparities in extracting location information from humanitarian documents. We introduce a two-step framework that combines few-shot LLM-based named entity recognition with an agent-based geocoding module that leverages context to resolve ambiguous toponyms. We benchmark our approach against state-of-the-art pretrained and rule-based systems using both accuracy and fairness metrics across geographic and socioeconomic dimensions. Our evaluation uses an extended version of the \textsc{HumSet} dataset with refined literal toponym annotations. Results show that LLM-based methods  substantially improve both the precision and fairness of geolocation extraction from humanitarian texts, particularly for underrepresented regions. By bridging advances in LLM reasoning with principles of responsible and inclusive AI, this work contributes to more equitable geospatial data systems for humanitarian response, advancing the goal of leaving no place behind in crisis analytics.
\end{abstract}

%%
%% The code below is generated by the tool at http://dl.acm.org/ccs.cfm.
%% Please copy and paste the code instead of the example below.
%%


%%
%% Keywords. The author(s) should pick words that accurately describe
%% the work being presented. Separate the keywords with commas.
%% A "teaser" image appears between the author and affiliation
%% information and the body of the document, and typically spans the
%% page.

%\received{20 February 2007}
%\received[revised]{12 March 2009}
%\received[accepted]{5 June 2009}

%%
%% This command processes the author and affiliation and title
%% information and builds the first part of the formatted document.

\maketitle

\section{Introduction}

Humanitarian crises, whether caused by natural disasters, conflicts, or pandemics, unfold across complex geographic, social, and linguistic landscapes. Timely and effective humanitarian response depends critically on knowing \textit{where} events occur. Despite the abundance of digital information available today, from situation reports to NGO bulletins and news coverage, automated systems still struggle to extract accurate and equitable geolocation data from text. These limitations not only hinder situational awareness but also risk perpetuating informational inequalities between regions that are richly represented in data and those that remain underreported.

Previous work has shown that existing geolocation pipelines systematically perform better for Western and wealthier countries, while underperforming in low and middle income regions. In particular, a recent study~\cite{leave_no_place_behind} evidenced that biases in both training data and gazetteers lead to an overrepresentation of locations from richer and English speaking regions in automatic geolocation outputs. Such disparities compromise the fairness of humanitarian analytics, where the accuracy of location information can directly affect the visibility of crisis affected communities.

In this work we investigate whether Large Language Models (LLMs) can mitigate these geographic disparities. We design a two step framework for fair geolocation extraction from humanitarian documents, combining a few shot LLM based NER tagger with an agent based geocoding module capable of leveraging context to resolve ambiguous or low resource toponyms. Our analysis focuses not only on overall performance but also on how and where LLMs make errors, quantifying improvements and residual biases across geographic and socioeconomic dimensions.

We benchmark our models against state of the art pretrained and rule based systems, using both accuracy and fairness metrics, and introduce an improved version of the \textsc{HumSet} dataset with refined literal toponym annotations. By bridging advances in large language models with the social mission of humanitarian data analysis, this work aims to leave no place behind, evaluating whether LLMs can produce more inclusive, transparent, and responsible geolocation systems for global crisis response.

\section{Related Work}~\label{sec:related}

Named Entity Recognition (NER) plays a central role in understanding news and reports about humanitarian crises, enabling the extraction of entities such as people, organizations, and locations related to ongoing events~\cite{nlp_humcrises_survey}. Traditional NER models, such as those based on \textsc{SpaCy} and \textsc{RoBERTa}, have achieved strong performance on benchmark datasets. However, they often display inconsistencies in how they tag geographic entities~\cite{pragmatic_geoparsing}, which complicates the extraction of meaningful geographic information from humanitarian text. 

The work of \citet{leave_no_place_behind} highlighted this challenge by proposing a hybrid approach combining \textsc{SpaCy} and \textsc{RoBERTa} with transfer learning and manual disambiguation between literal and associative toponyms. Despite improved precision, such approaches require extensive manual annotation and domain-specific heuristics to merge entities and ensure correct tagging. These limitations are particularly evident in humanitarian documents, which are longer, multilingual, and contain complex toponymic expressions (e.g., \emph{``the corner of Lyons Ave \& Gregg St''}) that differ from the short, structured texts used in standard NER corpora.

Recent advances in Large Language Models (LLMs) have demonstrated remarkable zero- and few-shot capabilities across a variety of natural language tasks~\cite{zero_shot_wei, zero_shot_brown}. LLMs have been successfully applied to general-purpose information extraction~\cite{llm_extraction, llm_extraction2}, as well as in more specialised tasks such as cross-domain adaptation~\cite{instruction_ner}, open named entity recognition~\cite{universal_ner}, domain-specific NER~\cite{medical_ner_gpt}, context-dependent location extractions~\cite{optimized_rag_llm_ner} and russian NER~\cite{russian_llm_ner}. Two studies, GPT-NER~\cite{gpt_ner} and PromptNER~\cite{prompt_ner}, use LLM-based approaches achieve results comparable to supervised deep learning models in the general purpose NER CoNLL dataset, while avoiding the need for large annotated datasets.

Large Language Models are increasingly used in the humanitarian and disaster management domains, where they have been applied to information extraction, situation summarisation, and crisis mapping~\cite{llm_extraction2, llms_disasters2}. Studies show that LLMs outperform traditional NER models in extracting locations from disaster-related social media data~\cite{geo_knowledge_gpt,geo_ocr_gpt} while multimodal approaches have leveraged both text and imagery for identifying affected areas~\cite{llm_geo_multimodal}.
uses images and text to extract the locations of social media tweets that request assistance in an undergoing disaster. However these studies are focused on short text on social media, that contain few locations and with almost all toponyms related to the location of the disaster. One study~\cite{chinese_collapse_llm_ner} leveraged LLMs performance on larger texts for identifying urban collapse events from chinese news and extracting the spatio-temporal information.%\KK{why is this relevant? (i havent' read it)}

Bias in geospatial data systems remains an open problem. Datasets for NER tagging~\cite{twits_ner_bias}, gazetteers such as geonames~\cite{geonames_bias}, geocoding systems~\cite{geocoding_linguistics} and pretrained supervised models~\cite{geoparsing_biased, woman_is_location, ner_models_language_change, leave_no_place_behind} have been shown to exhibit geographic, socioeconomic, and linguistic disparities. 
Recent analyses reveal that even foundation models display geographic biases and distortions~\cite{llms_geospatial_challenges, llms_distortions, llms_are_geo_biased, unequal_llm_recommendations}, which can propagate into humanitarian decision-making pipelines. \citet{biased_llm_geoclocation} further demonstrate that such biases extend to geolocation reasoning itself.

In related work on fairness in NER, \citet{ner_person_bias1, ner_person_bias2} show that performance disparities may arise based on name gender, ethnicity, or popularity, while \citet{bias_ner_location} identify differences between urban and rural toponym recognition. Yet, to our knowledge, a  systematically evaluation of fairness in location-based NER or geocoding with state-of-the-art LLMs is missing, and this is exactly the gap this paper aims to address.

Once location entities are extracted, geocoding methods link them to geographic coordinates or administrative boundaries. Earlier approaches relied on rule-based systems~\cite{leave_no_place_behind}, machine learning and deep learning based geocoding~\cite{entity_caillaut_2024}, gazetteer matching and hybrid approaches~\cite{geocoding_survey}. Recent transformer based architectures~\cite{transformers_toponym} have demonstrated significant improvements in resolving ambiguous toponyms~\cite{toponym_resolution}, offering context-aware reasoning capabilities unavailable to traditional systems. 

Building on this progress, our work introduces a fully LLM-driven two-step pipeline that combines NER and geocoding while evaluating fairness across geographic and economic strata, thereby bridging the gap between technical performance and equitable humanitarian applications.

\section{Methodology}~\label{sec:methodology}

\begin{figure*}[t]
  \centering
  \includegraphics[width=\textwidth]{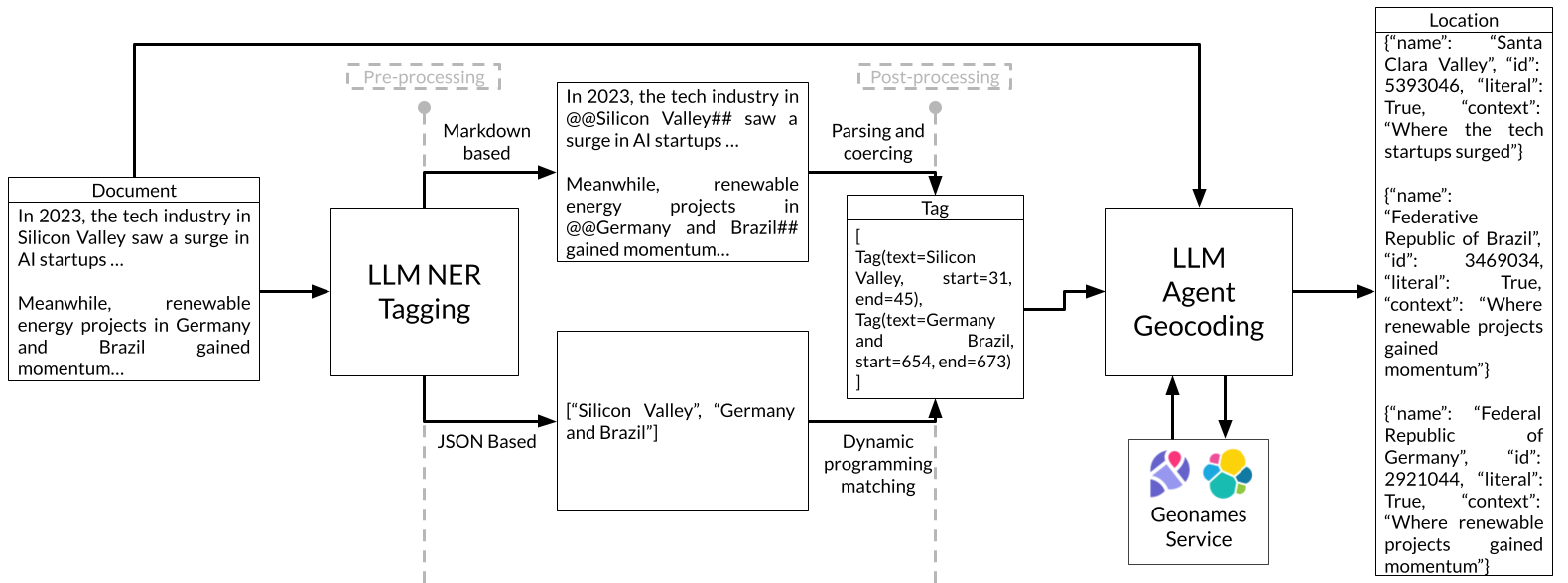}
\caption{Overview of our LLM-based geolocation extraction pipeline. Humanitarian documents are chunked and sent to a few-shot LLM for location NER tagging, followed by postprocessing that aligns and merges literal toponyms. An agent-based geocoder then uses contextual reasoning and GeoNames queries to resolve each toponym to coordinates, enabling us to evaluate both accuracy and fairness of geolocation across geographic and socioeconomic groups.}
  \label{fig:architecture}
\end{figure*}

We propose a pipeline for geolocation extraction based on LLMs illustrated in Figure~\ref{fig:architecture}. It consists of four main steps: document preprocessing, NER tagging, output postprocessing, and agent-based geocoding, which we next describe.

\paragraph{Document preprocessing}\label{sec:chunking} 
LLMs may struggle to retain long texts verbatim and to preserve the original order of information. To address this, we employed a chunking strategy based on the following rules:
(i) each chunk must fall within a predefined minimum and maximum length depending on the output format; (ii) documents are split into chunks using various candidate separators;
(iii) we consider the following candidate split separators, ordered from most preferred to least: double line break, ``. '', single line break, tabulation, ``, '', white space;
(iv) separators are tested sequentially until one produces chunks whose lengths fall within the target length range;
(v) once a suitable separator is found, the split configuration that minimizes the chunk length variance is chosen.
The specific minimum and maximum chunk lengths depend on the tagging format requested from the LLM, as detailed in the following subsection.

\paragraph{LLM-based NER tagging}\label{sec:ner} 
We propose a few-shot approach for NER tagging based on LLMs. 
We kept the same input data for all the LLM-based models we tested. Since the LLMs exhibit performance differences depending on the output format~\cite{output_format1, output_format2, output_format3} we evaluate two distinct formats: one based on JSON and another one based on Markdown. The choice of JSON output enables the LLM to process larger text segments without requiring verbatim reproduction of the entire input. However, a limitation is that the positions of the toponyms within the original text must be determined in a subsequent post-processing step.
The markdown approach on the other side, has the  advantage of directly providing the position of each toponym within the text, eliminating the need for the post-processing step required in the JSON-based approach. However, it is more error-prone because LLMs struggle to reproduce text verbatim as length increases.
The requested format will be specified in the prompt, after describing the task to the LLM, and before providing the example. The designed prompt is shown in Appendix~\ref{app:prompt_ner}.

In the JSON-based approach, the LLM is instructed to return a plain list of location strings in JSON format (e.g., \texttt{[``Milan'', ``Naples'', ``Rome'']}), where each element represents a literal toponym extracted verbatim from the input text. 
For our Markdown-based approach we follow the output structure proposed by~\citet{gpt_ner}, which reproduces the full text delimiting each toponym by \verb|@@| and \verb|##|. Given the size limitation of this approach, we reduce the chunk size to between 200 and 500 characters, compared to the JSON-based approach (1000--2000 characters), to improve reliability.
 
\paragraph{Post-processing}\label{sec:post_processing}
For the JSON-based approach, the LLM returns a list of toponyms without positional information. A greedy approach that processes each toponym sequentially and matches it to the first available occurrence after the previous match would introduce errors if the tags are misordered or have extra duplicates (e.g., in the phrase ``After evacuating from Goma, aid convoys moved through Bukavu, continued to Kigali, and ultimately returned to Goma,'' if the list is [``Goma'', ``Goma'', ``Bukavu'', ``Kigali''], the greedy algorithm will only recognize the first two appearances of ``Goma''). To avoid these mistakes, we use a dynamic-programming alignment that, for each toponym, considers all valid occurrences after the current index (including the option to skip a toponym) and selects the globally optimal, order-preserving assignment that maximizes the number of matched toponyms. As a limitation, this approach cannot handle cases in which a place is mentioned multiple times, appearing sometimes as a literal toponym and other times as an associative one (e.g., in ``The US Embassy in Nairobi, working with diplomats and local NGOs in the US, coordinated the humanitarian aid,'' the second appearance of ``US'' is the one we should match).

The Markdown-based approach generally requires only verifying that the tagged output chunk matches the input chunk, then extracting the delimited tags and their positions. However, when the output does not exactly match the input, we fall back to treating the extracted toponyms as a list and apply the same dynamic programming alignment algorithm used in the JSON-based approach.

Finally, and since we wanted all references to locations within the same phrase to appear as a single tag (e.g., ''Milan, Naples, and Rome''), in both approaches we merged nearby matches that are separated by characters as `,' or connectors like `and'. This post-processing step was necessary because the LLM sometimes failed to group them correctly or because the chunking process split them across segments. For a fair comparison, we also applied this correction during the re-annotation of the dataset and in the baseline models we compared against.

\paragraph{Geolocation Agent}\label{sec:agent} Once tags are extracted, we process each together with its surrounding document context using a geolocation agent. This agent was designed to identify all individual toponyms referenced within the tag and associate each one with its corresponding location. We implemented our agent in LangChain~\cite{langchain} and used the GeoNames database as the primary source of geographic information~\cite{geonames}. We used the open-source geocoding service Pelias~\cite{pelias} for querying GeoNames.

For each tag the agent alternates between reasoning steps and tool calls until task completion. We define our agent by providing a task description (see prompt in Appendix~\ref{app:prompt_agent}) and configuring the following steps: (i) \textit{Search:} a string in geonames matching a text and optionally a country code.
(ii) \textit{Select:} a geoname id for a place. The parameters are the place extracted from the text by the LLM (since multiple places may be in the same tag), the geoname id, a context added by the llm (e.g., ``Buenos Aires is where the earthquake happened'') and a flag indicating if the toponym is literal to further improve the toponyms dataset.
(iii) \textit{Finish:} the agent, with a reason to end the geocoding explaining why the task is done. This is used as a way to get some interpretability and insights over the reasoning of the LLM.

It is worth mentioning that the LLM agent inevitably inherits the structural biases present in GeoNames. By iteratively combining contextual inference with targeted search actions, the agent is able to resolve ambiguous or low-resource toponyms, distinguish between literal and associative place mentions, and justify its selections through explicit reasoning traces. This yields a more transparent and interpretable geocoding process, one that not only improves accuracy but also helps surface and mitigate geographic disparities in downstream predictions.

\subsection{Dataset}

For the evaluation and comparison of our models we used the \textsc{Humset}~\cite{humset} dataset which consists of $15,661$ multilingual reports, news and official documents about humanitarian response operations.
From them, we use the 467 English documents annotated by the previous study~\cite{leave_no_place_behind}; this enables us to perform a direct comparison with their NER model based on \textsc{SpaCy} and \textsc{RoBERTa} and their rule-based geocoding system. %The dataset consists of $15,661$ multilingual documents from 45 humanitarian response emergencies;
The decision to choose documents in English is due to the fact that international agencies that distribute these reports (e.g, ReliefWeb) publish primarily in this language, which serves as the lingua franca for international humanitarian collaboration.

We build upon the classification of toponyms into literal and associative categories established in prior research~\cite{pragmatic_geoparsing}. Literal toponyms refer to specific geographic locations, such as ``Buenos Aires'', while associative toponyms may refer to broader concepts or entities related to a location, like ``the Ambassador of Argentina''. In this work we focus on literal toponyms, as they are crucial for accurately identifying places linked to humanitarian crises.

\paragraph{Annotations' improvement} We found that some toponyms present in the dataset were missing or were incomplete in the annotations provided by~\citet{leave_no_place_behind}. Also, some tags that were annotated as literal toponyms were in fact associative. 
To address these issues, we developed a GUI tool that allowed our annotators to compare the output of two sets of NER tags and positions for the same document, pinpointing the differences between them and allowing the annotator to select the preferred ones. While this method is not as thorough as manually tagging from scratch, it enables a rapid improvement of the dataset.  Using this tool, the previously annotated dataset was compared to the output of our GPT 4o JSON model by a human annotator who solved all the conflicts. Then, the same process was repeated for this  new ground truth by a second human annotator who used the tool to compare it against the output of the \textsc{SpaCy} TRF model (the best of the pretrained classical models we tested). This resulted in an annotation set containing $8,213$ toponyms. 
Additionally, we requested an additional output to the geocoding agent asking it to flag those toponyms that were not actually literal. This allowed us to improve the dataset further: $1,534$ toponyms were flagged by the geolocator as not literal; these were also reviewed by two human annotators and resulted in removing $205$ toponyms from the final annotation set. Our improved set of annotations, along with the experimental code, are available at \url{https://github.com/jian01/llm-fair-geolocation-extraction-WWW2026}.

\subsection{Experiments}

With the improved NER annotation dataset we compared our proposed architecture to the state-of-the-art alternatives. We focused on two types of experiments: NER tagging performance and geocoding performance. For each of them we computed accuracy metrics such as precision, recall and F1 score, and fairness metrics. These experiments benchmark our LLM-driven approach against established methods, providing insights into the advancements made in NER and geolocation tasks through the integration of large language models.

\paragraph{NER tagging performance}
We assessed the performance of different language models for NER tagging under both JSON and Markdown output configurations. We evaluated GPT 5, GPT 4o, GPT 4o mini, Deepseek (reasoning and chat versions), Claude Sonnet 4.5, and Claude Haiku 4.5. We also evaluated a zero-shot approach proposed by PromptNER~\cite{prompt_ner} as implemented in the \textsc{SpaCy}-LLM ~\cite{spacy_llm} library using GPT 4o. 
We compare the performance of our LLM-based NER taggers against multiple baselines. We include the pre-trained models used in \citet{leave_no_place_behind} (\textsc{SpaCy} MD and XLM-RoBERTa) and add the transformer variant \textsc{SpaCy} TRF as being closer to the state of the art. Additionally, to assess the potential benefits of transfer learning, we fine-tune XLM-RoBERTa and \textsc{SpaCy} TRF on our improved annotation set using 5-fold cross-validation.
While \citet{leave_no_place_behind} proposed an ensemble of fine-tuned \textsc{RoBERTa} and \textsc{SpaCy} models, we do not include this combined variant in our comparison. In our setting, fine-tuned \textsc{RoBERTa} performs substantially below \textsc{SpaCy} TRF, making the ensemble redundant for assessing the relative benefits of LLM-based taggers.
In all baselines we used the same strategy for merging same-phrase locations as in our LLM-based NER-taggers to ensure a fair comparison.

Since some toponyms were captured partially (e.g. ``Rosario'' and ``Santa Fe province'' instead of ``Rosario in Santa Fe province''), we took precision, recall and F1 score metrics both considering exact matches (i.e., requiring the model to capture the full length of the toponym for the tag to be correct), and partial matches, where any tag found inside other tag in the ground truth is considered a true positive, and only tags that have no intersection with the ones in the dataset are considered false positives. 
By combining exact and partial matching metrics we obtain a more  comprehensive view of model performance: exact matching enforces strict adherence to our domain-specific annotation guidelines useful in the context of humanitarian geolocation, while partial matching provides a more lenient evaluation of literal toponym identification, offering insights into the model's core ability to detect relevant locations in a recall-oriented setting. Importantly, in our target deployment scenario, documents with highlighted locations will be reviewed by domain experts. In this human-in-the-loop workflow, recall is more critical than precision: false positives can be easily identified and dismissed by experts during review, whereas false negatives (missed locations) may go entirely unnoticed, potentially omitting critical geographic information from crisis response decisions.

\paragraph{Geolocation performance.}
For the geocoding task, we compared our LLM-based geolocation agent against the rule-based geocoding algorithm proposed by \citet{leave_no_place_behind}.
Using a sample of 10\% of the documents (47 documents containing a total of 969 toponyms) we geolocated the toponyms using the LLM agent and the rule-based algorithm proposed in~\citet{leave_no_place_behind}. The outputs of each model were reviewed by two independent annotators. Then both of them reconciled differences on their criteria. The output is a set of geoname ids for each toponym, that we use as our gold standard for evaluating the geocoding performance of each method.
We assessed the geocoding performance of each method using the following metrics: accuracy by exact id (i.e., matching not only the coordinates but also the administrative region level), the 161km accuracy based on the distance between coordinates associated to the ids, and the country accuracy (i.e., matching the country associated to the ids).

\paragraph{Fairness Assessment.}
We assessed the fairness of our system by quantifying biases in both the NER tagging models and the geolocators. Here we understand bias as the unequal performance of a model across different geographic and socioeconomic segments. Guided by equality of opportunity principles, we focus on measuring disparities in error rates across different groups rather than overall accuracy.
During the NER task, we consider a prediction correct if the predicted and ground-truth tags partially overlap, and we measure both the false negative rate (FNR) and the false discovery rate (FDR). For geographic segmentation, we group toponyms by continent (Africa, Asia, Europe, Americas). For socioeconomic segmentation, we use country income levels (low, lower-middle, upper-middle, high) based on the World Bank GNI classification~\cite{world_bank}. FNR captures the proportion of true location mentions that the model fails to identify high values in a region indicate that critical toponyms are systematically overlooked in that area. FDR captures the proportion of predicted locations that are incorrect high values indicate that the model over-identifies toponyms or misclassifies associative mentions (e.g., institutions) as literal places.

To quantify bias across groups, we use the equality-of-opportunity difference, denoted $\Delta\mathrm{FNR}$ and $\Delta\mathrm{FDR}$, defined as the difference between the highest and lowest error rates across all geographic or socioeconomic segments. The $\Delta$ value of ``0'' indicates perfect parity (equal performance across groups), whereas larger values reflect greater disparity.
For the geolocator, we consider a prediction correct if the predicted location lies within 161 km (approximately 100 miles) of our gold standard, a commonly used threshold in geolocation tasks~\cite{10040233}. We then compute the corresponding classification error across the same geographic and socioeconomic segments.

\section{Results}~\label{sec:results}

We evaluate our models along three axes: NER tagging, geolocation accuracy, and fairness across geographic and socioeconomic groups.

\begin{table*}[h!]
\centering
\begin{tabular}{lcccccccccc}
\toprule
\multicolumn{1}{c}{} & \multicolumn{4}{c}{\textbf{LLM-based models}} & \multicolumn{3}{c}{\textbf{Pre-trained models}} & \multicolumn{2}{c}{\textbf{Fine-tuned models}} \\
\cmidrule(lr){2-5} \cmidrule(lr){6-8} \cmidrule(lr){9-10}
 & \shortstack{GPT 5\\MD} & \shortstack{GPT 4o\\JSON} & \shortstack{GPT 4o\\MD} &  \shortstack{GPT 4o\\(zero shot)} & \shortstack{\textsc{\footnotesize{SpaCy}}\\MD} & \shortstack{\textsc{\footnotesize{RoBERTa}}\\vanilla} & 
 \shortstack{\textsc{\footnotesize{SpaCy}}\\TRF} & \shortstack{\textsc{\footnotesize{RoBERTa}}\\tuned} & \shortstack{\textsc{\footnotesize{SpaCy}} \\TRF tuned} \\
\midrule
\addlinespace[3pt]
\multicolumn{8}{l}{\textbf{Accuracy Metrics} \textit{(higher is better)}} \\
\addlinespace[3pt]
\;\textit{Exact} \\
\;\;Precision & 0.73 & \textbf{0.87} & 0.66 & 0.74 & 0.59 & 0.62 & 0.68 & 0.76 & 0.83 \\
\;\;Recall    & 0.83 & 0.82 & 0.84 & 0.80 & 0.52 & 0.27 & 0.70 & 0.44 & \textbf{0.85} \\
\;\;F1 Score        & 0.78 & \textbf{0.84} & 0.74 & 0.77 & 0.55 & 0.37 & 0.69 & 0.56 & \textbf{0.84} \\
\addlinespace[3pt]
\;\textit{Partial} \\
\;\;Precision & 0.87 & 0.90 & 0.75 & 0.83 & 0.87 & 0.90 & 0.92 & 0.90 & \textbf{0.93} \\
\;\;Recall    & \textbf{0.99} & 0.86 & 0.96 & 0.90 & 0.76 & 0.39 & 0.94 & 0.52 & 0.95 \\
\;\;F1 Score      & 0.92 & 0.88 & 0.84 & 0.87 & 0.81 & 0.54 & 0.93 & 0.66 & \textbf{0.94} \\
\hline
\addlinespace[3pt]
\multicolumn{8}{l}{\textbf{Fairness Metrics} \textit{(lower is better)}} \\
\addlinespace[3pt]
\multicolumn{8}{l}{\;\textit{Partial False Negative Rate by Continent}}\\
\;\;Asia     & \textbf{0.01} & 0.15 & 0.03 & 0.09 & 0.28 & 0.67 & 0.06 & 0.50 & 0.04 \\
\;\;Africa   & \textbf{0.01} & 0.14 & 0.03 & 0.09 & 0.25 & 0.60 & 0.06 & 0.46 & 0.04 \\
\;\;Americas & \textbf{0.01} & 0.10 & 0.03 & 0.08 & 0.11 & 0.49 & 0.02 & 0.44 & 0.03 \\
\;\;Europe   & 0.02 & 0.13 & 0.01 & 0.35 & 0.06 & 0.56 & \textbf{0.00} & 0.73 & 0.09 \\
\;\;$\Delta$FNR     & \textbf{0.01} & 0.05 & 0.02 & 0.27 & 0.22 & 0.17 & 0.06 & 0.29 & 0.06 \\

\addlinespace[3pt]
\multicolumn{8}{l}{\;\textit{Partial False Negative Rate by Country income level}}\\
\;\;Low Income            & \textbf{0.01} & 0.14 & 0.03 & 0.09 & 0.26 & 0.64 & 0.05 & 0.50 & 0.05 \\
\;\;Lower-middle income   & \textbf{0.02} & 0.12 & 0.03 & 0.09 & 0.26 & 0.58 & 0.09 & 0.42 & 0.05 \\
\;\;Upper-middle income   & \textbf{0.01} & 0.13 & 0.03 & 0.09 & 0.21 & 0.57 & 0.04 & 0.45 & 0.02 \\
\;\;High income           & \textbf{0.02} & 0.10 & 0.04 & 0.17 & 0.07 & 0.59 & 0.02 & 0.60 & 0.03 \\
\;\;$\Delta$FNR    & \textbf{0.01} & 0.04 & \textbf{0.01} & 0.08 & 0.18 & 0.07 & 0.08 & 0.18 & 0.03 \\

\addlinespace[3pt]
\multicolumn{8}{l}{\;\textit{Partial False Discovery Rate by Continent}}\\
\;\;Asia     & 0.21 & 0.11 & 0.29 & 0.19 & 0.13 & \textbf{0.09} & 0.12 & 0.09 & 0.09 \\
\;\;Africa   & 0.17 & 0.10 & 0.27 & 0.17 & 0.16 & 0.13 & \textbf{0.07} & 0.12 & 0.08 \\
\;\;Americas & 0.22 & 0.17 & 0.38 & 0.24 & 0.20 & \textbf{0.15} & 0.15 & 0.17 & 0.12 \\
\;\;Europe   & 0.73 & 0.54 & 0.75 & 0.74 & 0.79 & \textbf{0.53} & 0.71 & 0.61 & 0.63 \\
\;\;$\Delta$FDR     & 0.56 & 0.44 & 0.48 & 0.58 & 0.66 & \textbf{0.43} & 0.65 & 0.52 & 0.55 \\

\addlinespace[3pt]
\multicolumn{8}{l}{\;\textit{Partial False Discovery Rate by Country income level}}\\
\;\;Low Income            & 0.18 & 0.10 & 0.26 & 0.17 & 0.14 & 0.13 & 0.09 & 0.12 & \textbf{0.07} \\
\;\;Lower-middle income   & 0.23 & 0.11 & 0.31 & 0.19 & 0.14 & 0.10 & \textbf{0.09} & 0.11 & 0.08 \\
\;\;Upper-middle income   & 0.17 & 0.14 & 0.30 & 0.19 & 0.15 & 0.12 & \textbf{0.09} & 0.11 & 0.12 \\
\;\;High income           & 0.50 & 0.36 & 0.59 & 0.51 & 0.61 & \textbf{0.35} & 0.51 & 0.35 & 0.39 \\
\;\;$\Delta$FDR    & 0.33 & 0.26 & 0.33 & 0.34 & 0.47 & 0.25 & 0.42 & \textbf{0.24} & 0.32 \\

\bottomrule
\end{tabular}
\caption{Accuracy and fairness metrics for selected LLM-based NER taggers, compared against pretrained and fine-tuned deep learning baselines. The table reports exact and partial precision, recall, and F1 scores, as well as fairness metrics based on false negative and false discovery rates across geographic and socioeconomic groups. Higher accuracy values and lower disparity values indicate better performance.}
\label{tab:ner_metrics}
\end{table*}

\paragraph{\textbf{NER Tagging Performance}}
Table~\ref{tab:ner_metrics} reports accuracy and fairness metrics for a selection of the best-performing LLM-based NER taggers, alongside pretrained and fine-tuned baselines. The full set of accuracy metrics for all LLM-based models is provided in Appendix~\ref{app:all_acc}. 
Overall, Large Language Models outperform traditional pretrained deep learning models in both exact and partial NER tagging, without requiring task-specific training and while allowing flexible control over output format and toponym granularity. Fine-tuned models such as \textsc{SpaCy} TRF tuned achieve competitive scores relative to the best LLMs, but they depend on substantial annotated data and retraining when moving to new domains.

Among the LLM-based models, GPT-4o JSON attains the highest exact precision (0.87) and shares the best exact F1 score (0.84) with the fine-tuned \textsc{SpaCy} TRF model, making it a strong choice when precise extraction of literal toponyms is critical. \textsc{SpaCy} TRF tuned, in turn, reaches the highest exact recall (0.85), illustrating how fine-tuned architectures can be adapted to favor more enriched toponym spans (e.g., ``Buenos Aires city'' instead of ``Buenos Aires'').

For partial metrics, which emphasise the correct identification of literal toponyms even when span boundaries are not exact, GPT-5 Markdown achieves the highest recall (0.99). The best partial precision (0.93) and partial F1 score (0.94) are obtained by the fine-tuned \textsc{SpaCy} TRF model, which also already performs strongly in its pretrained form. By contrast, both pretrained and fine-tuned \textsc{RoBERTa} models exhibit consistently low recall (ranging from 0.27 to 0.55), making them unreliable for capturing the full set of relevant locations in humanitarian texts. Finally, the zero-shot GPT-4o model attains F1 scores that are comparable to, and in some cases higher than, other LLM configurations, underscoring the ability of modern LLMs to deliver competitive performance even without task-specific examples.

\begin{figure}[htbp]
    \centering
    \includegraphics[width=\linewidth]{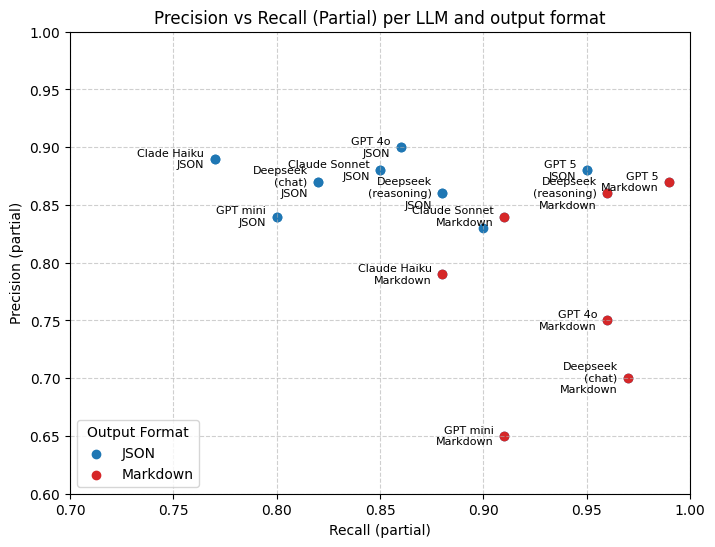}
    \caption{Precision–recall trade-offs for the evaluated LLM-based NER taggers under JSON and Markdown output formats. JSON configurations generally yield higher precision, while Markdown formats achieve higher recall, reflecting the tendency of Markdown prompts to capture more toponyms at the cost of increased false positives.}
    \label{fig:prec_recall_format}
\end{figure}

Regarding output formatting, we observe systematic differences between JSON and Markdown configurations, as shown in Figure~\ref{fig:prec_recall_format}. JSON-based prompts tend to achieve higher precision, whereas Markdown-based prompts yield higher recall. One plausible explanation is that Markdown outputs must reproduce the input text verbatim, which may make it easier not to miss location mentions, but also more likely to introduce false positives, particularly for associative toponyms. These format-dependent trade-offs highlight the importance of prompt and output design when deploying LLMs for location extraction in humanitarian text.

\begin{figure*}[t!]
  \centering
  \includegraphics[width=\textwidth]{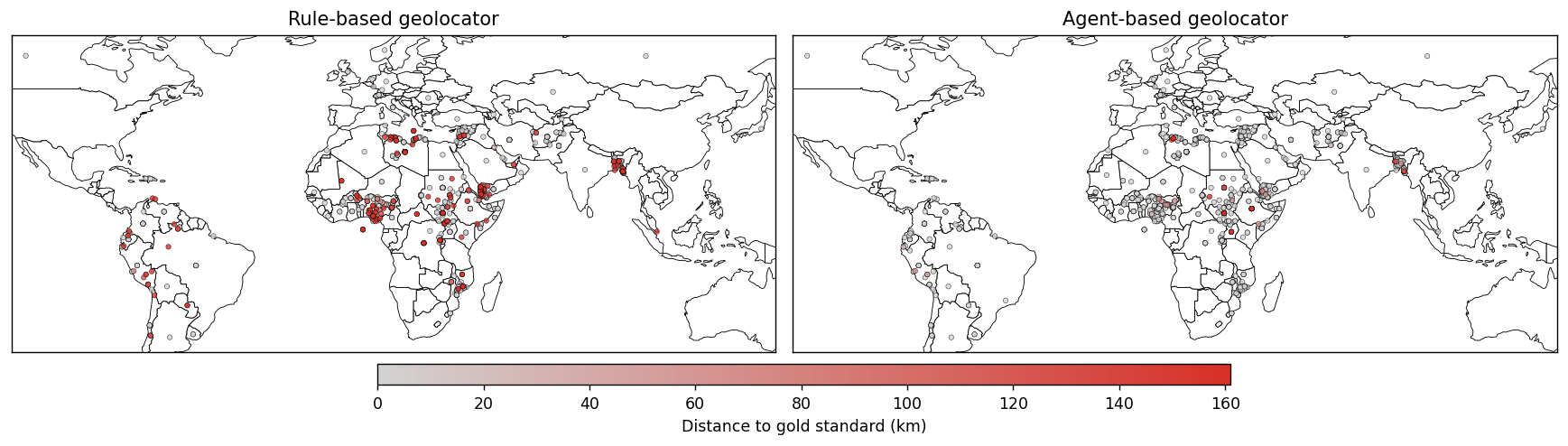}
 \caption{Geographic distribution of geocoding errors for the rule-based baseline (left) and our agent-based geolocator (right). Each point represents a toponym from the annotated evaluation set, plotted at its gold-standard location and colored by distance (in km) between the system’s predicted coordinates and the ground truth. Darker colors indicate larger errors. The rule-based method exhibits frequent long-distance misclassifications, including cross-continent mismatches, whereas the agent-based geolocator substantially reduces large-error cases and achieves more geographically consistent performance across regions.}
  \label{fig:errors_map}
\end{figure*}

\paragraph{\textbf{Geolocation Performance}}
Table~\ref{tab:geoc_metrics} summarizes the geocoding accuracy and fairness metrics for our LLM-based geolocation agent compared to the rule-based baseline from \citet{leave_no_place_behind}. The agent-based approach significantly outperforms the rule-based system across all accuracy metrics: exact, 161\,km, and country-level precision and recall are consistently higher for the agent-based model.
Figure~\ref{fig:errors_map} visualizes the spatial distribution of prediction errors for both geocoding approaches, measured as the distance between each model's predicted location and our gold-standard annotations. The maps show that our agent-based geolocator achieves substantially lower errors across all world regions compared to the rule-based baseline, reducing large-distance errors and cross-continent mismatches.

\begin{table}[h!]
\centering
\begin{tabular}{lcc}
\toprule
 & \shortstack{Agent-based} & \shortstack{Rule-based~\cite{leave_no_place_behind}} \\
\midrule
\addlinespace[3pt]
\multicolumn{3}{l}{\textbf{Accuracy Metrics} \textit{(higher is better)}} \\
\addlinespace[3pt]
\textit{Exact} \\
\;Precision      & \textbf{0.88} & 0.69 \\
\;Recall      & \textbf{0.79} & 0.59 \\
\addlinespace[3pt]
\textit{@161km} \\
\;Precision      & \textbf{0.91} & 0.83 \\
\;Recall       & \textbf{0.82} & 0.71 \\
\addlinespace[3pt]
\textit{Country-level} \\
\;Precision      & \textbf{0.91} & 0.82 \\
\;Recall      & \textbf{0.83} & 0.70 \\
\addlinespace[3pt]
\hline
\addlinespace[3pt]

\multicolumn{3}{l}{\textbf{Fairness Metrics} \textit{(lower is better)}
} \\

\addlinespace[3pt]
\multicolumn{3}{l}{\textit{Classif. error@161km by Continent}}\\
\;Asia     & \textbf{0.24} & 0.43 \\
\;Africa   & \textbf{0.25} & 0.34 \\
\;Americas & \textbf{0.06} & 0.44 \\
\;Europe & \textbf{0.00} & 0.58 \\
\;$\Delta$Error & 0.25 & \textbf{0.24} \\

\addlinespace[3pt]
\multicolumn{3}{l}{\textit{Classif. error@161km by Country income level}}\\
\;Low income           & \textbf{0.25} & 0.30 \\
\;Lower-middle income  & \textbf{0.21} & 0.44 \\
\;Upper-middle income  & \textbf{0.23} & 0.42 \\
\;High income  & \textbf{0.07} & 0.65 \\
\;$\Delta$Error & \textbf{0.18} & 0.35 \\
\bottomrule
\end{tabular}
\caption{Accuracy and fairness metrics for the geocoding task, comparing our agent-based LLM geolocator with the rule-based baseline from \citet{leave_no_place_behind}. Higher accuracy and lower disparity indicate better performance. The agent-based model achieves substantial improvements in both accuracy and geographic parity.}
\label{tab:geoc_metrics}
\end{table}

Table~\ref{tab:geoc_error_examples} illustrates representative prediction errors for each geocoding approach. In the first example, \texttt{eastern South Sudan}, the agent-based model incorrectly resolves to a more specific administrative region (Eastern Equatoria) rather than the country-level entity, likely because the LLM over-interprets the directional modifier ``eastern'' as requiring a sub-national match. The second example, \texttt{central Mediterranean}, showcases a failure unique to the rule-based approach: it incorrectly matches to Davao in the Philippines, possibly due to gazetteer ambiguity or lack of contextual reasoning, while the agent successfully identifies the Mediterranean Sea. Notably, one of the alternative names for Davao in GeoNames is ``central'', which may have confused the rule-based model. Finally, both models fail on \texttt{Dar-fur}: the agent selects North Darfur (a sub-region) instead of the broader Darfur region, while the rule-based system erroneously links it to Dar es Salaam in Tanzania, likely misled by partial string matching and the hyphen symbol. These examples highlight that while the agent-based approach generally outperforms the baseline, it can still over-specialise when strong contextual cues are present, whereas rule-based methods remain vulnerable to superficial lexical similarities.

\paragraph{\textbf{Fairness Assessment}}
We now analyse how performance varies across geographic and socioeconomic segments, focusing on disparities in error rates rather than aggregate accuracy. For NER tagging, we consider a prediction correct if the predicted and ground-truth tags partially overlap, and measure false negative rates (FNR) and false discovery rates (FDR) across continents (Africa, Asia, Europe, Americas) and World Bank income groups (low, lower-middle, upper-middle, high~\cite{world_bank}). To summarise disparities, we report equality-of-opportunity differences $\Delta\mathrm{FNR}$ and $\Delta\mathrm{FDR}$, defined as the difference between the highest and lowest error rates across segments (0 indicates perfect parity, larger values indicate greater disparity).

In terms of FNR, Table~\ref{tab:ner_metrics} shows that GPT-5 Markdown and GPT-4o Markdown maintain near-parity across continents and income levels, with maximum differences of around 1 percentage point between groups. This suggests that their extraction ability remains stable across geographic and socioeconomic contexts, mitigating the regional biases observed in earlier systems. In contrast, \textsc{RoBERTa} variants exhibit stronger disparities, while \textsc{SpaCy} TRF achieves relatively low FDR but lags behind in exact tagging accuracy. The zero-shot GPT-4o configuration displays among the highest disparities overall, possibly because the limited in-context examples underrepresent certain regions and thus shift the model’s implicit bias. Taking both accuracy and fairness into account, GPT-5 Markdown emerges as the best overall NER model, achieving nearly uniform FNR across groups while maintaining high partial recall (0.99) and F1 (0.92).

FDR remains a challenge for all models, with disparities of up to 0.66 driven largely by high false positive rates in Europe. The particularly low proportion of true literal toponyms located in Europe (below $2\%$) makes its FDR highly sensitive to model misclassifications.

For the geocoding task, we consider a prediction correct if the predicted location lies within 161\,km (approximately 100 miles) of our gold standard~\cite{10040233}, and compute the corresponding classification error@161\,km across the same segments, summarising disparities with $\Delta\mathrm{Error}$.

As shown in Table~\ref{tab:geoc_metrics}, our agent-based geolocator reduces classification error@161\,km in all continents and income groups relative to the rule-based baseline. The rule-based system exhibits particularly high error in Europe and high-income countries, reflecting frequent cross-country or cross-continent mismatches. The agent-based approach not only lowers overall error but also reduces $\Delta\mathrm{Error}$ by roughly half when stratified by income level, indicating more uniform performance across economic strata and making it better suited for equitable geolocation in humanitarian contexts.

\begin{table*}[h!]
\centering
\begin{tabular}{lcccc}
\toprule
 & \textbf{Toponym} &
\textbf{Gold Standard} &
\textbf{Agent-based} &
\textbf{Rule-based} \\
\midrule
Agent-only & \texttt{eastern South Sudan} &
\shortstack{South Sudan} &
\shortstack{Eastern Equatoria} &
\shortstack{South Sudan} \\

\addlinespace[5pt]
Rule-only & \texttt{central Mediterranean} &
\shortstack{Mediterranean Sea} &
\shortstack{Mediterranean Sea} &
\shortstack{Davao} \\

\addlinespace[5pt]
Both & \texttt{Dar-fur} &
\shortstack{Darfur} &
\shortstack{North Darfur} &
\shortstack{Dar es Salaam} \\
\bottomrule
\end{tabular}
\caption{Representative geocoding error cases illustrating distinct failure modes of the agent-based and rule-based geolocators. Each row shows a toponym from the evaluation set, its gold-standard location, and the predictions from both systems. The examples highlight agent-only over-specification (\texttt{eastern South Sudan}), rule-based lexical mismatches (\texttt{central Mediterranean}), and ambiguous or nonstandard toponyms that challenge both methods (\texttt{Dar-fur}).}

\label{tab:geoc_error_examples}
\end{table*}

\section{Discussion}~\label{sec:discussion}

Building on the findings of \emph{Leave No Place Behind}~\cite{leave_no_place_behind}, our results confirm that improving the quality and representativeness of training data remains central to mitigating geographic bias in geolocation extraction. 
While earlier work showed that tuning classical NER models such as \textsc{SpaCy} and \textsc{RoBERTa} on humanitarian data improved accuracy and slightly alleviated Western bias, our study demonstrates that Large Language Models can further advance both objectives. 

While \textsc{SpaCy} TRF tuned achieves metrics comparable to the best LLM-based models, it is important to note that our LLM configurations operate in a few-shot regime, enabling the development of domain-specific NER taggers with minimal annotation effort. Future work should investigate whether supervised prompt optimisation~\cite{dspy} can further improve few-shot performance and potentially surpass the gains achieved through classical fine-tuning.

Our results also show that NER performance with LLMs depends strongly on the chosen output format. Formatting thus becomes an integral part of prompt design and requires deliberate selection and tuning. Markdown-based outputs, in particular, tend to generate more tokens, which can increase computational cost and should be taken into account when selecting a production configuration.

For NER tagging, both \textsc{GPT-5 Markdown} and \textsc{GPT-4o Markdown} display near-parity in false negative rates across continents and income groups, indicating that their extraction ability is geographically and socioeconomically balanced. In contrast, pretrained and fine-tuned \textsc{SpaCy} TRF models exhibit larger disparities and tend to underperform in the Global South, particularly in lower-income regions. For geocoding, our agent-based approach substantially improves almost all accuracy and fairness metrics across groups compared to the rule-based baseline.

The progression from domain-tuned transformers~\cite{leave_no_place_behind} to foundation LLMs marks a substantial methodological shift: these models no longer require explicit retraining to adapt to humanitarian text, yet they deliver superior results with less manual intervention. 
This opens new opportunities for rapid deployment in multilingual and low-resource humanitarian contexts, where labeled data are scarce. 
Nevertheless, as our analysis shows, even the most advanced LLMs are not free from residual geographic and socioeconomic biases, emphasizing the importance of continuous algorithmic auditing and inclusion of diverse linguistic and regional data sources.

Overall, our results demonstrate that LLM-driven geolocation not only improves extraction quality, but also meaningfully reduces geographic disparities, helping build more inclusive and reliable information pipelines for humanitarian response.
Compared to previous efforts that improved bias mitigation through fine-tuning~\cite{leave_no_place_behind}, LLMs can achieve comparable or better fairness out-of-the-box, while also advancing the state of the art in extraction accuracy, enabling adaptive systems that can further reduce bias and 
ensure that, truly, \emph{no place is left behind}.

\section{Acknowledgments}

KK and YM acknowledge support from the Lagrange Project of the Institute for Scientific Interchange Foundation (ISI Foundation) funded by Fondazione Cassa di Risparmio di Torino (Fondazione CRT). MGB and GC acknowledge financial support from Universidad de San Andrés.

\newpage

%%% -*-BibTeX-*-
%%% Do NOT edit. File created by BibTeX with style
%%% ACM-Reference-Format-Journals [18-Jan-2012].

\newpage

\appendix

\section{NER tagging prompt}\label{app:prompt_ner}

Table~\ref{fig:prompt_ner} shows the prompt structure used for the NER tagging task.\\

\begin{table}[h]
\centering\scriptsize
\begin{verbatim}
You are a Named Entity Recognition (NER) system specialized in extracting
**literal toponyms** (geographic location names) from texts about natural 
disasters and accidents. 
Your task is to identify and return **only the explicit literal mentions of 
physical locations** (toponyms), avoiding any associative uses.
Each time the user gives you a text you simply answer each occurence of an 
explicit literal toponym avoiding associative toponyms.

Key constraints:
- Extract only literal toponyms, defined as:
  - Proper names of places (e.g., "Cambridge", "Germany")
  - Noun modifiers of places (e.g., "Paris pub")
  - Adjectival modifiers with geographic meaning (e.g., "southern Spanish 
  city")

- Do NOT extract associative toponyms, including:
  - Metonymic references (e.g., "She used to play for Cambridge")
  - Demonyms (e.g., "a Jamaican")
  - Homonyms (e.g., "I asked Paris to help")
  - Languages (e.g., "in Spanish")
  - Noun/adjectival modifiers not referring to a physical place (e.g., 
  "Spanish ham")
  - Embedded uses (e.g., "US Supreme Court", "US Dollar")
  - Toponyms in URLs

Extraction Rules:
1. Preserve literal mentions exactly as they appear — no rephrasing or 
normalization.
2. Preserve order — output the locations in the same order as in the input.
3. Do not remove duplicates.
4. Include:
   - Geographical regions (e.g., "Patagonia", "coastal Germany")
   - Roads, borders, and composite names (e.g., "Buenos Aires–Mar del Plata 
   road")
   - Temporary places like refugee camps
   - Institutions only if they imply a geographic location (e.g., "Cambridge 
   University" implies "Cambridge")
5. Prefer the most specific geographic level available (e.g., "Buenos Aires 
province" over "Buenos Aires").
6. Include articles if they are part of the place name.
7. Include **cardinal directions** (e.g., "southern Spain").
8. **Do not merge** multiple toponyms unless they jointly modify a noun 
immediately after (e.g., "South Dakota, New York and Michigan states" → 
merge all three).
9. If multiple locations are listed and are connected by commas, “and”, or 
“in” within a single continuous phrase, keep them together as one literal 
toponym exactly as 
written, even if they contain multiple place names.
10. Merge nested location phrases with possessive or relational structure
\end{verbatim}
\caption{Prompt used for the LLM-based NER taggers.}
\label{fig:prompt_ner}
\end{table}

\newpage

\section{Geolocator agent prompt}\label{app:prompt_agent}

Table~\ref{fig:prompt_agent} shows the prompt structure used for the geolocator agent.
\vspace{-0.5cm}
\begin{table}[h]
\centering\scriptsize
\begin{verbatim}
## Task & Tools

You work as a **geolocator**. A location (PLACE) and its **context**
will be provided. Your goal is to maximize 100 mile accuracy. Use the
**tools**
1) `search_tool` candidates in GeoNames database. You can pass an
ISO country code when the context hints at it (e.g., 'AR', 'BR', 'MZ'),
2) For each **definite** location, call **`select_tool`** (you may
call it multiple times) for select the best matching entry. Be precise
and prefer cities/settlements over vague regions when the
context points to them,
3) When you are done, call **`finish_tool`** exactly once to
return **all selections together**.

## Considerations

- If PLACE is ambiguous, refine with `search_tool` (try alternate spellings
from context).
- When PLACE is a country name, you MUST select the sovereign state entry
(feature_code PCL*).
- If a PCL* candidate is not returned on the first search, try alternate
spellings or exonyms/endonyms and search again.
- If the context indicates a country explicitly, include it in searches.
- Prefer inhabited places (feature classes `P*`) when context refers to
education, health, local news, municipal services, etc.
- If the text clearly references a **province/state/county** instead of a
city, select that administrative division.
- Avoid false positives from homonyms in other countries—check language,
nearby mentions, and timezone cues in context.
- If no good candidate appears, perform another `search_tool` with a better
query (e.g., remove accents, try shorter stems).
- If PLACE matches a sovereign state (country name), prefer the country
entry (feature_code = 'PCLI') rather than a city.
- Do not call `select_tool` twice for the same geonameid or the same place.
- You must cover every sublocation enumerated in PLACE; do not call
finish_tool until each is either selected or explicitly marked with
select_tool(geonameid=-1, ...).
- You only have 15 actions to use.
- Whether relevant or not, try to locate all toponyms, only the ones not in
geonames should be left without selection.
- Prefer shorter searches and do not insist too much on one location, you
have limited action, prioritize selection before searching.
- Do at most **only two** searches per place.
- Do not extract implicit location from within the context or not explicitly
named or directly associated to the place string
- If the toponym is "<X> in <Y>" extract just <X> unless <X> is not in
geonames, then extract <Y>
- Do **not** add extra locations not related to the place input

## Non-literal (associative) toponyms

Examples of non-literal toponyms:
* Metonymy: She used to play for **Cambridge**.
* Homonym: I asked **Paris** to help me.
* Demonym: I spoke to a **Jamaican** on the bus.
* Language: She spoked **Spanish**.
* Noum modifier: That **Paris** souvenir is interesting.
* Adjectival modifier: I ate some **Spanish** ham yesterday.
* Embedded associative: **US** Supreme Court has 9 justices.
You should mark those kind of toponyms as not literal.

## Output protocol
- Call `select_tool(geonameid, context, literal_toponym)` for every
confirmed location.
  - geonameid should **always** be present as a valid geonames id, get the
closest one possible.
  - `context` is an English explanation like "Neighborhood mentioned in Yola
North LGA", or
    "Government jurisdiction", etc. If the toponym is not literal you should
specify why.
  - `literal_toponym` is True when the toponym refers to a real place
related to what the document is describing, False when associative.
  - You should select only one id per place and then finish.
- Finally, call `finish_tool(reason=...)` once. Do not return free-form text.
\end{verbatim}
\caption{Prompt used for the geocoding agent.}
\label{fig:prompt_agent}
\end{table}

\section{Accuracy metrics for all LLM-based NER taggers and output formats}\label{app:all_acc}

Table~\ref{tab:all_models_metrics} shows the accuracy metrics for all the LLM-based models we assessed for the NER tagging task.

\begin{table}[h]
\centering
\begin{tabular}{l *{6}{c}}  % lcccccc but shorter
\toprule
\multicolumn{1}{c}{} &
\multicolumn{3}{c}{\textbf{Exact}} &
\multicolumn{3}{c}{\textbf{Partial}} \\
\cmidrule(lr){2-4} \cmidrule(lr){5-7}
Model &
Prec. & Rec. & F1 &
Prec. & Rec. & F1 \\
\midrule

\textit{GPT 5}\\
\;\;JSON            & 0.76 & 0.82 & 0.79 & 0.88 & 0.95 & \textbf{0.92} \\
\;\;MD              & 0.73 & 0.83 & 0.78 & 0.87 & \textbf{0.99} & \textbf{0.92} \\

\textit{GPT 4o}\\
\;\;JSON           & \textbf{0.87} & 0.82 & \textbf{0.84} & \textbf{0.90} & 0.86 & 0.88 \\
\;\;MD             & 0.66 & \textbf{0.84} & 0.74 & 0.75 & 0.96 & 0.84 \\
\;\;zero-shot & 0.74 & 0.80 & 0.77 & 0.83 & 0.90 & 0.87 \\

\textit{GPT mini}\\
\;\;JSON         & 0.74 & 0.71 & 0.72 & 0.84 & 0.80 & 0.82 \\
\;\;MD           & 0.56 & 0.79 & 0.66 & 0.65 & 0.91 & 0.76 \\

\textit{Deepseek reason.}\\
\;\;JSON & 0.77 & 0.79 & 0.78 & 0.86 & 0.88 & 0.87 \\
\;\;MD   & 0.74 & 0.83 & 0.78 & 0.86 & 0.96 & 0.91 \\

\textit{Deepseek chat}\\
\;\;JSON     & 0.78 & 0.73 & 0.75 & 0.87 & 0.82 & 0.84 \\
\;\;MD       & 0.59 & 0.81 & 0.68 & 0.70 & 0.97 & 0.81 \\

\textit{Claude Sonnet}\\
\;\;JSON     & 0.79 & 0.76 & 0.77 & 0.88 & 0.85 & 0.87 \\
\;\;MD       & 0.73 & 0.79 & 0.76 & 0.84 & 0.91 & 0.88 \\

\textit{Claude Haiku}\\
\;\;JSON      & 0.79 & 0.69 & 0.73 & 0.89 & 0.77 & 0.82 \\
\;\;MD        & 0.69 & 0.76 & 0.72 & 0.79 & 0.88 & 0.83 \\

\bottomrule
\end{tabular}
\caption{Accuracy metrics for all LLM-based models for the NER tagging task. Bold numbers indicate the best model for each metric.}
\label{tab:all_models_metrics}
\end{table}

\end{document}